\documentclass[10pt,letterpaper]{article}

\usepackage{cogsci}

\cogscifinalcopy 

\usepackage{pslatex}
\usepackage{apacite}
\usepackage{float} 

\usepackage{tikz-dependency}
\usepackage{verbatim}
\usepackage{amsmath}
\usepackage{epsfig}
\usepackage{multirow}
\usepackage{linguex}
\usepackage{graphicx}
\usepackage{subfig}
\usepackage{devanagari}
\usepackage{wrapfig}
\usepackage{cases}
\usepackage{color}
\usepackage{booktabs}
\usepackage{nameref}
\usepackage[titletoc]{appendix}
\usepackage{mathtools}
\usepackage{lipsum}  
\usepackage{comment}
\usepackage{xurl}
\usepackage{commath}
\usepackage{gb4e}

\title{Does Dependency Locality Predict Non-canonical Word Order in Hindi?}
 
\author{{\large \bf Sidharth Ranjan (sidharth.ranjan@ling.uni-stuttgart.de)} \\
 University of Stuttgart, 70174 Stuttgart, Germany  \\
  \AND {\large \bf Marten van Schijndel (mv443@cornell.edu)} \\
  Cornell University, 14853 Ithaca, United States}

\begin{document}

\maketitle

\begin{abstract}

    Previous work has shown that isolated non-canonical sentences with {Object-before-Subject} (OSV) order are initially harder to process than their canonical counterparts with {Subject-before-Object} (SOV) order. Although this difficulty diminishes with appropriate discourse context, the underlying cognitive factors responsible for alleviating processing challenges in OSV sentences remain a question. In this work, we test the hypothesis that dependency length minimization is a significant predictor of non-canonical (OSV) syntactic choices, especially when controlling for information status such as givenness and surprisal measures. We extract sentences from the Hindi-Urdu Treebank corpus (HUTB) that contain clearly-defined subjects and objects, systematically permute the preverbal constituents of those sentences, and deploy a classifier to distinguish between original corpus sentences and artificially generated alternatives. The classifier leverages various discourse-based and cognitive features, including dependency length, surprisal, and information status, to inform its predictions. Our results suggest that, although there exists a preference for minimizing dependency length in non-canonical corpus sentences amidst the generated variants, this factor does not significantly contribute in identifying corpus sentences above and beyond surprisal and givenness measures. Notably, discourse predictability emerges as the primary determinant of constituent-order preferences. These findings are further supported by human evaluations involving 44 native Hindi speakers. Overall, this work sheds light on the role of expectation adaptation in word-ordering decisions. We conclude by situating our results within the theories of discourse production and information locality.

\end{abstract}

\section{Introduction}\label{sect:intro}

The \textsc{subject (S), object (O),} and \textsc{verb (V)} collectively constitute the building blocks or constituents of a sentence, operating together in a structured manner to convey the intended meaning. However, the linguistic properties of each language result in a spectrum of legitimate constituent orders within the sentence which all potentially express the same idea~\cite{chomsky1965aspects,ferreira1996better}, some of these are more common than others. The most frequent constituent order is often called \emph{canonical} and is often considered the most natural and basic syntactic structure in a language. It is typically considered neutral and unmarked. In contrast, the alternative \emph{non-canonical} orders are rarer and are employed under specific information structural and discourse considerations~\cite{comrie1981,gambhirphd,Bock1982,mohanan1994wordorder}. Comprehenders often have greater difficulty understanding these less frequent constructions than those that are commonly produced~\cite{hale2001,levy2008,hawkins2014}. However, the same rare non-canonical orders, when used within an appropriate discourse or situational context, can be as preferable and easy to process as sentences with canonical orders~\cite{kaiser2004role,hornig2005two,kristensen2014context,bader2023processing}. 


While Hindi, a flexible word order language~\cite{kachru1982}, usually employs SOV order, \citeA{vasishth:jsal2012} found that context rendered certain non-canonical word orders easier to comprehend than their canonical counterparts, suggesting that syntactic complexity can be mitigated by context. In spite of this, \citeA{vasishthysall04} has shown that some non-canonical OSV sentences such as \ref{ex:hindi-ref} remain harder to comprehend than \ref{ex:hindi-var} despite the presence of an appropriate discourse context (\ref{ex:hindi-context}). This difficulty seems to arise due to a considerably larger distance between the fronted object and the verb, thereby increasing the total dependency length within sentence~\cite{gibson00}.

\begin{small}
\begin{exe}

  \ex 
  {\label{ex:hindi-context} \textbf{Discourse context sentence}\footnote{Major constituents are annotated with brackets, repeated content words are underlined in both context and target sentences.}
    \gll {air india} {aur} {indian} {{\color{blue}{{\underline{vimaan}}}} sevaayon-ko} {\color{blue}\underline{{notice}}} {de} {diya gya hai}\\
         {Air India} {and} {Indian} {airlines services-\textsc{acc}} {notice} {give} {give\textsc{.prs.pfv.sg}}\\
     \glt \textit{Air India and Indian Airlines services have been notified.}}

\end{exe}

\begin{exe}

  \ex \label{ex:hindi-intro} \textbf{Target sentence}
  \begin{xlist}
  \ex[]{\label{ex:hindi-ref}
    \gll {[yah {\color{blue}\underline{notice}}]\color{red}{$_{DO}$}} {[kuwait-ke} {\color{blue}{\underline{vimaan}}} {pradhikaran} {mantralaye-ke} {director general-ne]\color{red}{$_{S}$}} {jaari} {[kiya]\color{red}{$_V$}} {hai} \textbf{(Reference; OSV;  Given-Given)}\\
         {this notice} {kuwait-\textsc{gen}} {aviation} {authority} {ministry-\textsc{gen}} {general-\textsc{erg}} {issue} {do} {be\textsc{.prs.pfv.sg}}\\
     \glt {\it This notice has been issued by the Director-General of the Ministry of Aviation Ministry of Kuwait}.}
  \ex[] {\label{ex:hindi-var} {[kuwait-ke} {{\color{blue}\underline{vimaan}} pradhikaran} {mantralaye-ke} {director general-ne]\color{red}{$_{S}$}} {[yah {\color{blue}\underline{notice}}]\color{red}{$_{DO}$}} {jaari} {[kiya]\color{red}{$_V$}} {hai} \textbf{(Variant; SOV; Given-Given)}}
  \end{xlist}

\end{exe}
\end{small}

In this work, we study the cognitive factors that influence a speaker's choice to produce a non-canonical order when an equally viable canonical order exists in Hindi. While naturally occurring non-canonical sentences do minimize dependency length as compared with the canonical alternatives, we find that givenness and surprisal minimization are stronger predictors of this behavior.

\section{Hypothesis}

In a recent study, \citeA{cog:sid} demonstrated that although dependency length minimization is generally a weak predictor of Hindi word order preferences, it significantly predicts non-canonical (OSV) orders, even in presence of surprisal estimates from trigram and incremental dependency parsing models. They argued that surprisal measures may perform poorly for non-canonical orders since the models for estimating surprisal are themselves biased towards more frequent structures, while dependency length minimization proved effective because fronted-objects were consistently longer than the subjects, resulting in sequences favoring long-before-short patterns in the corpus. 

Following their work, we set out to test the hypothesis that dependency length minimization is a significant predictor of non-canonical (OSV) production choices, especially when controlling for information status such as givenness and surprisal. 
To do this we use dependency length, givenness, and a variety of surprisal measures to predict the likelihood of non-canonical vs canonical constructions in naturally occurring written sentences from the Hindi-Urdu Treebank corpus~(HUTB; \citeNP{Bhatt2009}).
If a factor influences production decisions in humans, it should be able to correctly anticipate where different constituent orderings will occur in the corpus.

\section{Measures of Processing Difficulty}\label{sect:bkg}

\subsection{Dependency Locality Theory}

Dependency Locality Theory~(DLT; \citeNP{Gib98,gibson00}) has proven effective both in predicting comprehension difficulty and in modeling production durations~\cite{scontras2014,dammalapati2021expectation}. 

DLT posits two types of processing costs: \textsc{integration cost} and \textsc{storage cost}.  
In general, shorter dependencies have been found to dominate across natural languages~\cite{Liu2008,futrell2020dependency,RanjanMalsburg2023CogSci,RanjanMalsburg2024CogSci}. The placement of \textit{long-before-short} constituents in SOV languages and \textit{short-before-long} constituents in SVO languages is preferred as they minimize total dependency length within the sentence~\cite{hawkins1994,hawkins04}. In this study, we define dependency length as the count of intervening words between head and dependent units within a dependency graph, and calculate the sentence-level dependency length by summing each word's dependency length within the sentence~\cite{Temperley2007}.

\subsection{Surprisal}

Claude Shannon's \citeyear{shannon1948} work on Information Theory defined the amount of new information contained in an observation, based on the (un)predictability of that observation before it occurred. \citeA{hale2001} showed that this measure, called \emph{surprisal}, can predict language processing behavior:

\begin{small}
\begin{equation}{\label{eq2}}
S_{k+1} = -\log P(w_{k+1}|w_{1...k},~C)
\end{equation}
\end{small}

The \textit{surprisal} of a word ($w_{k+1}$) is defined in terms of the conditional probability of that word, $w$ given its intra-sentential context ($w_{1...k}$) and inter-sentential context ($C$; Eq. \ref{eq2}). These probabilities can be estimated using either a \textit{sequence of words} or a \textit{syntactic tree} \cite{levy2008}. Many studies have now shown that words with high surprisal are difficult to process and comprehend \cite{jurafsky1996probabilistic,hale2006uncertainty,DembergKeller2008,smith-levy:2013}. 
In this work, we estimate per-word surprisal using a number of methods described below and then compute the sentence-level surprisal for each method by summing the per-word scores within each sentence.

\begin{enumerate}
    \item \textbf{$n$-gram surprisal:} To capture linear local predictability, we used a traditional trigram language model to estimate the surprisal of the target word conditioned only on the two preceding context words.
    The trigram model we used was trained on 1 million sentences from the EMILLE corpus~\cite{emille2002} with Good-Turing discounting using the SRILM toolkit~\cite{SRILM-ICSLP:2002}.
    
    \item \textbf{PCFG surprisal:} To capture structural predictability, we computed the conditional probability of the words given the incremental available syntactic structures. This metric was computed for each word in the sentence using the Modelblocks incremental constituency parser~\cite{vanschijndeletal13:topics}.\footnote{\url{https://github.com/modelblocks}} 
    Because there are no constituency treebanks for Hindi, we used the~\cite{Yadav2017KeepingIS} method to generate 12,000 constituency trees from the dependencies in the HUTB corpus. Since we wanted to use these trees both to train our parser and to analyze in our study, we used a 5-fold cross-validation (CV) method whereby each fifth of our data was annotated with PCFG surprisal after training our parser on the other four-fifths of our data.\footnote{F1-score: 90.82\% (train) and 84.95\% (test) in 5-fold CV setting}

    \item \textbf{Adaptive LSTM surprisal:} To capture long-range predictability effects, we used the Neural Complexity Toolkit~\cite{van2018neural}.\footnote{\url{https://github.com/vansky/neural-complexity}} This toolkit implements online fine-tuning of a neural language model, where an LSTM language model is updated on each context sentence to better model in influence of discourse context (Example \ref{ex:hindi-context}), and then surprisal of the words in the target sentences (Example \ref{ex:hindi-intro}) are estimated using the revised language model weights. This approach generates discourse-sensitive surprisal estimates that considers not only intra-sentential unbounded context but also inter-sentential information. Using this method, \citeA{van2018neural} found that adaptive surprisal was significantly better at predicting human reading times than non-adaptive surprisal. 

\end{enumerate}

\subsection{Information Status}

Languages are known to follow a principle of  \textsc{given-before-new} ordering, by re-referring to information previously mentioned in the discourse \textit{before} introducing the new information in the sentence~\cite{Clark-Haviland77}. 
 
We annotate each sentence in our dataset with Information Status (IS). For this, we examine subject and object phrases in the target sentences (\ref{ex:hindi-intro}), checking if any content words within these phrases are mentioned in the preceding context sentence (\ref{ex:hindi-context}). If so, these phrases are tagged as \textsc{given}, otherwise they are tagged as \textsc{new}. Additionally, we label phrases as \textsc{given} if the head of these constituents is a pronoun. 
Under this scheme, both reference (\ref{ex:hindi-ref}) and variant (\ref{ex:hindi-var}) sentences display a \textsc{given-given} ordering as both the subject and the object contain content words (underlined) that are present in the context sentence (\ref{ex:hindi-context}). Numerical IS scores were assigned as follows: a) Given-New order = +1, b) New-Given order = -1, c) Given-Given and New-New orders = 0.
\section{Data and Methods}\label{sect:method}

In our study, we used sentences from the HUTB corpus of written newswire sentences~\cite{Bhatt2009}. We only analyzed those trees that adhered to the following criteria: a) the trees contain both well-defined subjects and objects, b) the trees are projective, c) the sentences are declarative, and d) the root node for each tree is a finite verb with at least two preverbal dependents.

We generated counterfactual variants for each reference sentence in our dataset. This involved randomly permuting the preverbal constituents within the dependency tree, specifically those directly dependent on the root verb (see Ex.~\ref{ex:hindi-intro}). The preverbal domain serves as the primary locus of constituent order variation in SOV languages, with fewer post-verbal constituents~\cite{cog:sid}. We limited the variant generation to a maximum of 99 variants per corpus reference sentence as an arbitrary cutoff for computational tractability, though our findings remained consistent regardless of the chosen cutoff. Ungrammatical variants were automatically filtered out using grammar rules derived from the corpus dependency trees in the treebank~\cite{llc:raja:mwhite:2014}. 

\begin{table}
\centering
\scalebox{0.75}{
    \begin{tabular}{l|c|l|l}
        Dataset & Ordering type & Reference (\%) & Variant (\%) \\\hline
        O-fronted & Ref: OSV; Var: SOV & 233 (11.67\%) & 2834 (3.89\%)\\
        DO-fronted & Ref: DOSV; Var: SDOV & 133 (6.67\%) & 1663 (2.28\%)\\
        IO-fronted & Ref: IOSV; Var: SIOV & 101 (5.06\%) & 1353 (1.86\%)\\ \hline
        Canonical & Ref: SOV; Var: OSV & 1763 (88.33\%) & 30851 (42.36\%) \\\hline
        \multicolumn{2}{c|}{Full dataset} & 1996 & 72833\\
    \end{tabular}}
    \vspace{-0.7em}
    \caption{Distribution of ordering types in Reference (Ref) and Variant (Var) sentences in our dataset; S = subject; O = object; V = verb; DO = direct object; IO = indirect object}
    \label{tab:raw:dataset}
\end{table}

Our variant generation procedure resulted in a total of 72833 grammatical variants for 1996 reference sentences in the corpus. We then categorized reference and variants by their word orders into canonical (SOV) and non-canonical (OSV) types, as illustrated in Table~\ref{tab:raw:dataset}. The occurrence distribution revealed that non-canonical reference sentences are indeed less frequent compared to canonical orders. Following this, we set up a classification task to investigate if the features described in the preceding section are effective in identifying the corpus reference sentences amidst competing counterfactual variants. However, the presence of a greater number of variants (72833) per reference sentence (1996) introduces a substantial class imbalance issue ({reference} / {variant}) for our binary classification task. The ranking method in the next section
addresses this class imbalance problem. We then use the transformed dataset with balanced class labels to report our regression and classification analyses.

We also conducted a targeted human evaluation via a two-alternative forced choice (2AFC) judgment task involving 44 Hindi native speakers. We collected human judgments over 164 randomly chosen reference-variant pairs from our full dataset. Participants first viewed the context sentence (Ex.~\ref{ex:hindi-context}) and then selected the preferred continuation between the reference (Ex.~\ref{ex:hindi-ref}) and the variant (Ex.~\ref{ex:hindi-var}). 
Results showed that, 85.63\% of the HUTB reference sentences (out of 164 pairs) were preferred by native speakers over the plausible variants that we generated. As a result, we feel justified in considering the reference sentences in the HUTB corpus as gold standards of human ordering preferences in our analyses. 

\subsection{Ranking Model}

Our original dataset contained significantly more variants (72833) than reference sentences (1996) and this imbalance posed a problem for training an unbiased classifier. To rectify this imbalance, we applied a data transformation technique originally proposed for ranking web pages by~\citeA{Joachims:2002}. Previous work on syntactic choice phenomena has successfully employed this method, which converts a binary classification task with heavily imbalanced labels into a pairwise ranking task~\cite{cog:raja}. This conversion associates feature vectors from the reference sentences with those of their variants. The feature vectors in this case consist of the aforementioned features which we computed for both reference and variant sentences. Next, we trained a logistic regression model on the difference between feature vectors of reference and variant pairs, as illustrated in the equations below:

\begin{small}
   \begin{equation}
\label{eq:refvar1}
\mathbf{w}\cdot\phi(Reference)>\mathbf{w}\cdot\phi(Variant)
   \end{equation}
\begin{equation}
\label{eq:refvar2}
\mathbf{w}\cdot\mathbf{(}\phi(Reference)-\phi(Variant)\mathbf{)}>0
\end{equation}
   
\end{small}
   
Eq.~\ref{eq:refvar1} depicts a standard classification model determining whether the reference sentence outperforms one of its variants. This decision relies on comparing the dot product of the feature vector linked to the reference sentence and the learned feature weights $\mathbf{w}$ with the corresponding dot product of the variant sentence. This relationship is alternatively expressed in Eq.~\ref{eq:refvar2}, where the feature values of the first pair member are subtracted from their counterparts in the second member~\cite{Joachims:2002}. The model's decision for a specific reference-variant pair can then be made by evaluating the sign of the dot product between the learned feature weights and the feature vector difference (Eq.~\ref{eq:refvar2}).

We generated ordered pairs comprising feature vectors for both reference (\textsc{ref}) and variant (\textsc{var}) sentences, maintaining a balance in the counts of each order type (\textsc{ref-var}, \textsc{var-ref}).
Pairs alternating between `\textsc{ref-var}' were labeled as `1,' whereas pairs in the `\textsc{var-ref}' sequence were labeled as `0.' This coding strategy ensures a balanced dataset, with a near equal number of labels for each type.\footnote{The number of labels is equal when the total number of variants is even and they differ by one when the variant total is odd}
The transformed features were then fed into a logistic regression model (using the \texttt{glm} function in R) to test our hypothesis:

\vspace{-1em}
\begin{small}
\begin{equation}\label{eq:regr}
choice \sim  \begin{cases}
 & \text{$\delta_0$ adaptive lstm surprisal +}\\
 & \text{$\delta_1$ trigram surprisal + $\delta_2$ pcfg surprisal +} \\ 
 & \text{$\delta_3$ IS score +} \\ 
  & \text{$\delta_4$ dependency length} 
\end{cases}
\end{equation}
\end{small}
\vspace{-0.5em}

Here, \textit{choice} is a binary dependent variable (1 denotes a preference for the reference sentence, and 0 denotes a preference for the variant sentence). The deltas ($\delta_x$) refer to the difference between each feature of the reference sentence and the paired variant. All the independent variables were normalised to $z$-scores.  

To evaluate the classification performance, we examined the output of our model trained on the entire dataset (72833 data points) with transformed feature values as predictors using 10-fold cross-validation method i.e., models trained on 9 folds of the dataset were used for prediction in the remaining fold. However, for estimating regression coefficients in the model, we use the test data with transformed feature values as predictors for a given construction under study.


\section{Results}\label{sect:exp}

In this section, we investigate whether dependency length minimization is a significant predictor of OSV syntactic choices, even in presence of surprisal and givenness measures. As discussed previously, corpus orderings were human-preferred compared with plausible alternatives. We assume that this is due to the psycholinguistic properties of each sentence. 
We first used regression to identify the strength with which each factor predicts the corpus orderings compared with possible but unattested alternatives. Then we examined the percentage of correctly predicted reference sentences over paired alternative variants using our features. Finally, we collected human preference data to determine how closely these features cause our automatic classifier to mimic human behavior.

\begin{figure}
    \centering
    \includegraphics[scale=0.2]{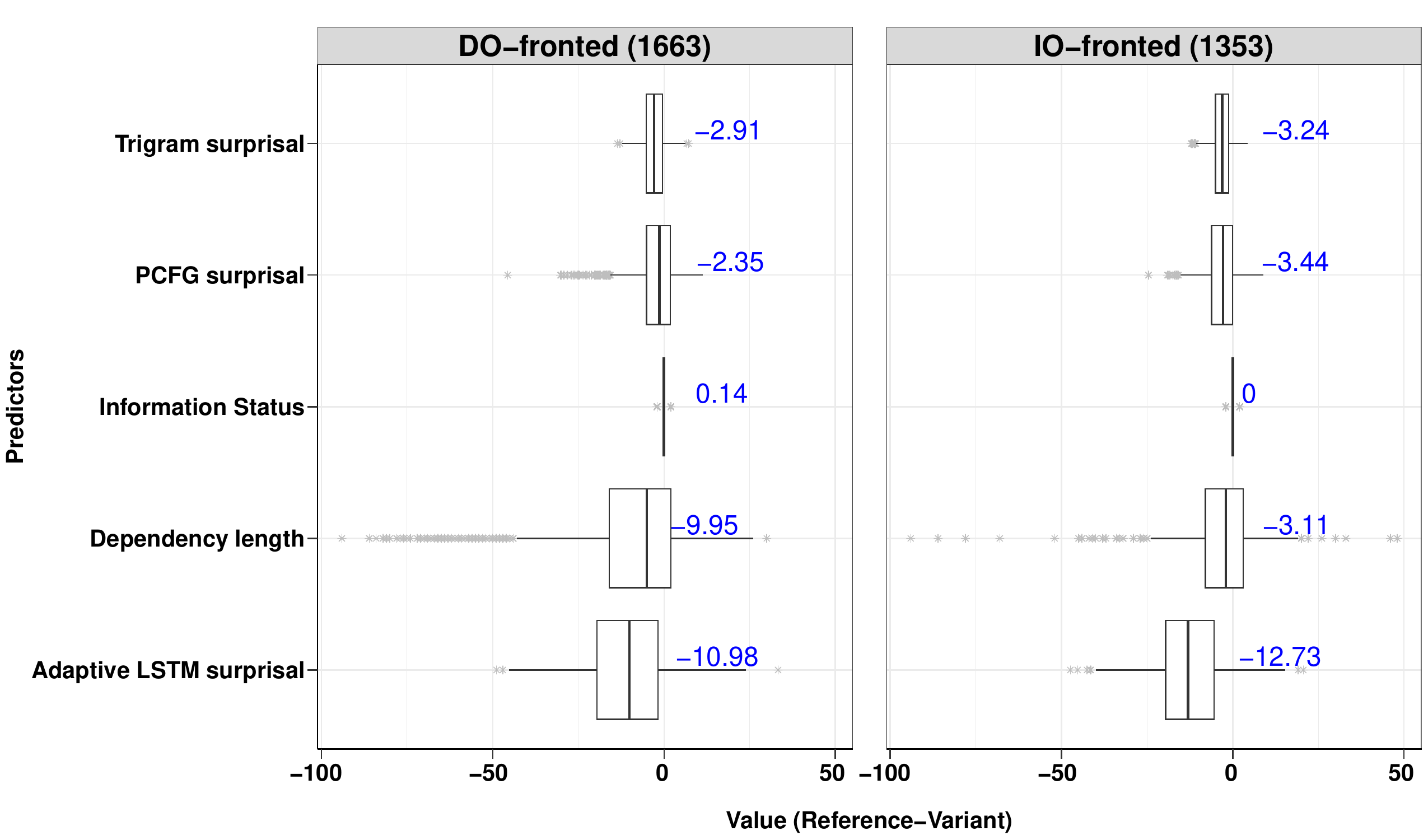}
    \vspace{-1.8em}
    \caption{Summary of dataset denoting difference between predictor values of Reference and their corresponding variant sentence for DO-fronted and IO-fronted constructions; Mean difference value for each predictor is annotated.}
    \label{fig:box-plot}
\end{figure}

\subsection{Regression Analysis}

\begin{table*}
\begin{minipage}{0.5\textwidth}
\centering

\scalebox{0.78}{
\begin{tabular}{l|cccc}
\textbf{Predictors} & \textbf{Estimate} & \textbf{Std. Error} & \textbf{Z-value} & \textbf{VIF} \\\hline

{Intercept} & 0.02 & 0.074 & 0.25 & --\\
{Dependency length} & \textbf{-0.58} & 0.091 & -6.35*** & 1.02\\
{IS score} & \textbf{0.26} & 0.055 & 4.68*** & 1.03\\
{PCFG surprisal} & \textbf{0.52} & 0.135 & 3.88*** & 1.42\\
{Trigram surprisal} & \textbf{-2.08} & 0.245 & -8.48*** & 2.12\\
{Adaptive LSTM surprisal} & \textbf{-1.65} & 0.240 & -6.86*** & 1.99\\\hline

\end{tabular}}
\vspace{-0.5em}
\caption{Regression coefficients for the model predicting reference sentence as DO-fronted non-canonical (DOSV) against variant as canonical (SDOV) ($N=1663$; all significant predictors denoted by *** $p$~\textless~0.001; VIF = Variance Inflation Factor)}\label{reg:DO-Rnoncano-Vcano} 

\end{minipage}%
\quad
\begin{minipage}{0.5\textwidth}
\centering

\scalebox{0.78}{
\begin{tabular}{l|cccc}
\textbf{Predictors} & \textbf{Estimate} & \textbf{Std. Error} & \textbf{Z-value} & \textbf{VIF} \\\hline

{Intercept} & 0.20 & 0.109 & 1.875 & --\\
{Dependency length} & {0.15} & 0.142 & 1.045 & 1.08\\
{IS score} & {0.13} & 0.074 & 1.767 & 1.10\\
{PCFG surprisal} & \textbf{-0.89} & 0.238 & -3.725*** & 1.08\\
{Trigram surprisal} & \textbf{-3.05} & 0.382 & -7.976*** & 1.34\\
{Adaptive LSTM surprisal} & \textbf{-2.83} & 0.365 & -7.78*** & 1.43\\\hline

\end{tabular}}
\vspace{-0.5em}
\caption{Regression coefficients for the model predicting reference sentence as IO-fronted non-canonical (IOSV) against variant as canonical (SIOV) ($N=1353$; all significant predictors denoted by *** $p$~\textless~0.001; VIF = Variance Inflation Factor)}\label{reg:IO-Rnoncano-Vcano}

\end{minipage}
\end{table*}

In this section, we test our main hypothesis that dependency length predicts non-canonical OSV orders (i.e.\ DO-fronted and IO-fronted corpus sentences) while controlling for surprisal measures and IS score. 
Based on the prior literature, we expect that surprisal should be lower in the reference than in the variant and that givenness should be adhered to more in the reference sentence than the variant.\footnote{Based on the coding of given-new as +1 and new-given as -1, we expect the reference-variant difference [(+1) - (-1) = 2] for IS score to be a positive value if givenness is truly adhered to.} If dependency length minimization plays a role in these ordering decisions, we would expect dependency length to be lower in the reference sentences compared with the variants. Our results (Figure~\ref{fig:box-plot}) suggest that surprisal and dependency length predictions generally hold across both types of non-canonical constructions. However, the givenness (information status) prediction holds true only for DO-fronted sentences. 

For the DO-fronted subset of our corpus (Table~\ref{reg:DO-Rnoncano-Vcano}), reference sentences consistently minimize dependency length in addition to trigram and adaptive LSTM surprisal as shown by their negative regression coefficients. However, the positive regression coefficient for PCFG surprisal suggests greater syntactic complexity associated with DO-fronted constructions. The positive regression coefficient for IS score implies that DO-fronted corpus sentences exhibit \textsc{given-new} ordering as compared to the paired alternative variant. Adding dependency length to a baseline model containing all other predictors significantly improved the fit of our model in the DO-fronted subset ($\chi^2$ = 51.61; p $<$ 0.001). In contrast, for IO-fronted constructions (Table~\ref{reg:IO-Rnoncano-Vcano}), we observe that reference sentences consistently minimize all surprisal measures as evinced by their negative regression coefficients. Dependency length and IS score were not significantly predictive of IO-fronted orderings.

Our regression results are largely in line with previous findings in the literature, which have found differential processing effects of dependency length for DO- and IO-fronted constructions in Hindi~\cite{vasishthysall04,cog:sid,ranjan-etal-2022-discourse}. Ranjan and colleagues, in their analysis of HUTB sentences found a preference for \textit{long-before-short} constituent placement in DO-fronted sentences in line with dependency length minimization (DLM). In contrast, they found no such influence of dependency length in IO-fronted sentences. In the present work, we find that while OSV sentences optimize for surprisal minimization generally, only DO-fronted sentences tend to minimize dependency length.

\subsection{Classification Analysis}

\begin{table*}
\centering
\scalebox{0.74}{
\begin{tabular}{cccc}
Predictor(s)       & (Ref: OSV \textit{vs} Var: SOV) & (Ref: DOSV \textit{vs} Var: SDOV) & (Ref: IOSV \textit{vs} Var: SIOV)\\ \hline
IS Score & 52.19~~~~ & 53.88~~~~ & 50.92~~~~~~\\
Dependency length   & 65.24*** & 68.49*** & 58.91***\\
PCFG surprisal & 67.22~~~~ & 59.05*** & 75.91***\\
Trigram surprisal   & 83.16*** & 78.95*** & 87.29***\\
Adaptive LSTM surprisal   & 83.84~~~~ & 79.98~~~~ & 88.32~~~~~\\\hline
Baseline = Adaptive LSTM + trigram surprisal & 85.18*** & 81.24**~ & 89.43*\\
Baseline + PCFG surprisal & 84.72~~~~ & 80.04*** & 89.95~~~\\
Baseline + PCFG surprisal + dependency length & 84.65~~~~ & 79.86~~~~ & 89.95~~~\\
All predictors      & 85.04~~~~ & 80.46*~~ & 90.02~~~\\\hline
\end{tabular}
}
\vspace{-0.7em}
\caption{Individual and collective prediction accuracies (Random accuracy = 50\%; *** $p<0.001$ ; ** $p<0.01$; * $p<0.05$ McNemar's two-tailed significance compared to model on previous row; Refer to Table \ref{tab:raw:dataset} for our dataset distribution)}
\label{tab:Rnoncano-Vcano-Pred-Acc}
\end{table*}

In this section, we examine the impact of each factor in classifying whether a sentence is a corpus reference sentence or a variant. We use 10-fold cross-validation of the entire dataset (72833 data points) to evaluate the classifier's performance. Table~\ref{tab:Rnoncano-Vcano-Pred-Acc} presents the prediction accuracy of our various models across three constructions \textit{viz.,} object-fronted (OSV), DO-fronted (DOSV), and IO-fronted (IOSV) in our dataset. 

In terms of individual performance (top block in Table \ref{tab:Rnoncano-Vcano-Pred-Acc}), adaptive LSTM and trigram surprisal achieved the highest classification accuracy across all three constructions. PCFG surprisal is the second best predictor followed by dependency length and IS score in decreasing order for OSV and IOSV constructions. The poor performance of PCFG surprisal in predicting DOSV sentences is mainly due to its bias towards frequent syntactic structures, in contrast to dependency length, which quantifies memory load within sentence. These findings suggest that a syntactically complex sentence can be less taxing if the processing is facilitated via DLM or probable word sequences in the sentence.

Over a baseline model containing adaptive LSTM and trigram surprisal measures (bottom block in Table \ref{tab:Rnoncano-Vcano-Pred-Acc}), none of the predictors---PCFG surprisal, dependency length, and IS score---induced any increase in the classification performance for OSV (second column) and IOSV constructions (last column). However, for DOSV construction, PCFG surprisal significantly helped classify reference sentences above and beyond the baseline model (p $<$ 0.001 using McNemar's two-tailed significance test). Dependency length failed to contribute beyond the model containing all the surprisal measures. Interestingly, IS score successfully predicted DO-fronted constructions beyond the model containing all predictors, implying that fronting direct objects is driven by considerations of givenness in discourse. In another experiment, we also tested whether dependency length significantly contributes beyond a model with only adaptive LSTM surprisal, but found no significant impact in predicting corpus sentences across the three constructions. This implies that the impact of DLM is entirely subsumed by adaptive LSTM surprisal.

In sum, our classification analyses involving 10-fold cross-validation, focused on \textit{prediction}, stand in contrast to our regression analyses, which involved \textit{fitting} the data. Our classification findings refute the hypothesis that DLM significantly predicts non-canonical ordering choices, once expectation-based factors and discourse considerations are accounted for. Instead, surprisal minimization, captured by inter- and intra-sentential language models like adaptive LSTM and trigram surprisal, emerges as the primary driver of syntactic ordering choices in Hindi. Our work extends beyond prior research that solely incorporates intra-sentential surprisal measures for modeling syntactic choices~\cite{cog:raja,ranjan-etal-2019-surprisal}, and aligns with the recent findings that have attested the role of discourse context in sentence processing~\cite{TingJaeger2012,van2018neural,ranjan-etal-2022-discourse}.

\subsection{Human Evaluation}

\begin{table}
    \scalebox{0.75}{
    \begin{tabular}{l|cc|c}
        Dataset Type (\#)            & Corpus labels    & Human labels & Human-Corpus\\
                                & \multicolumn{2}{c|}{{Prediction Accuracy (\%)}} & Agreement (\%)\\\hline
        Ref: OSV; Var: SOV (41) & 53.66     & 70.73 & 63.41  \\
        DO-fronted  (20)             & 45.00     & 70.00 & 55.00\\
        IO-fronted (21)              & 61.90     & 71.43 & 71.43\\\hline
        Full dataset (164)                & \textbf{76.65}     & \textbf{79.04} & 85.63\\\hline
    \end{tabular}}
\vspace{-0.7em}
\caption{Model's performance on original corpus labels and the human labels from 44 native Hindi speakers}
    \label{tab:human-prediction}
\end{table}

While our previous analyses relied entirely on corpus occurrences as proxies for human preference data, this section investigates how well our classification model simulates real human judgments from 44 native speakers. 
As described in the Methods section, we collected human judgments using 164 randomly chosen reference-variant pairs from our full dataset (72833). We employed 10-fold cross-validation on the full dataset to assess our classifier's performance on these 164 pairs, trained with all features. Subsequently, we compared the classifier's output with the real-human choices. Table~\ref{tab:human-prediction} reports our human evaluation outcomes.

Our classifier consistently demonstrated better accuracy in predicting the human-preferred constructions across all the items we tested 
than in predicting original corpus choices. The increased accuracy for human labels (70\%) over corpus labels (45\%) in DO-fronted constructions revealed that native speakers generally prefer canonical forms over non-canonical ones even though the non-canonical sentences minimized dependency length on average (see Figure~\ref{fig:box-plot} and Table~\ref{reg:DO-Rnoncano-Vcano}), and our model is better at picking up those instances. We also found that Hindi participants favored IO-fronted constructions (71.43\%) over DO-fronted constructions (55\%) as captured by the human-corpus agreement accuracy, which reflects the number of sentences where the corpus reference sentence was preferred by human raters. Further data analysis revealed that our model struggled to predict human labels when participants themselves disagree among each other. We observed significant disagreement among native speakers for non-canonical (OSV) orders, and in particular, for DO-fronted reference sentences, thus serving as a primary reason underlying the differing performance of our classifier in predicting human and corpus labels.
\section{Discussion}\label{sect:disc}

Dependency Length Minimization seems to be a general property of many natural languages and it does seem to significantly predict general word ordering in Hindi. However, our work indicates that discourse expectations, as captured by adaptive LSTM surprisal and givenness seem to be the primary influences on non-canonical sentence production. Notably, discourse-enhanced surprisal entirely subsumes the impact of dependency length minimization effects in predicting Hindi syntactic choices. It is possible that dependency length partially helps determine when non-canonical constructions occur, but our results indicate that at least some other factors are also at work. We suspect that other cognitive mechanisms underlying surprisal minimization include accessibility of multi-word sequences~\cite{realiChristiansen2007}, considerations of 
local coherence effects~\cite{tabor2004}, and discourse adaptation, incorporating both lexical and syntactic priming~\cite{gries2005,tooley2010syntactic,reitter2011,ranjan-priming}.

Previous studies in Hindi have noted a weak locality effect, affecting sentences that are either syntactically complex or that involve rare constructions~\cite{vasishthlewisLanguage06,cog:sid}. However, our present findings reveal that, with the inclusion of discourse factors, the locality effect ceases to be a general predictor, even in less common non-canonical sentences. The present findings, therefore, align with the idea in the literature that expectation effects are likely to be prevalent in SOV languages, whereas locality effects are observed in SVO languages~\cite{vasishth2010,cancho2015,cancho2017}

A recent study by \citeA{hahn2020}, involving an extensive analysis of 51 languages, demonstrated that dependency-length minimization could be an epiphenomenon, sufficiently accounted for by optimizing the general predictability and parseability of sentences without requiring any additional constraints. 
Similarly, \citeA{engelmann2019} noted that decay component (representative of locality) in \citeA{lewisvasishth:cogsci05}'s model did not improve fit in their simulation data across various dependency types. These findings possibly suggest that \textit{decay}, the underlying cognitive construct behind ``locality" may lack strong empirical support~\cite{ol-wm-2013,oberauer2014further,stone2020effect,jager2020}, although see \citeA{hardt2013decay}. This casts doubt on the cause of dependency length minimization being solely attributed to decay in memory, and proposes an alternative explanation where locality might be reducible to a more general memory interference effect~\cite{vasishth:jsal2012,scil:sid:interference}. Future work needs to investigate the role of interference while controlling for locality and surprisal in syntactic choice phenomena in Hindi.

Our results also have implications for the Information Locality Hypothesis~(ILH), which integrates lossy context surprisal and mutual information~\cite{futrell-2019-information,futrell2020}. ILH suggests that words predicting each other should exhibit locality effects, and processing difficulty is related to a word's expected log probability given a noisy memory representation. With this view, ILH, thus is thought to unify surprisal and dependency locality, and proposes a metric based on point-wise mutual information (PMI) between word pairs. Our work uses an LSTM language model with an unbounded context, which effectively captures and subsumes the effects of locality for syntactic choice. Thus offering some preliminary evidence for ILH. Nevertheless, further research is required to test the original formulation of ILH using PMI metric for syntactic choice in Hindi.

Overall, our results provide converging evidence that while speakers of Hindi, an SOV language, display a weak tendency to minimize dependency length, the preverbal constituent orderings are primarily shaped by surprisal minimization leveraging both intra- and inter-level context information.


\section{Acknowledgments}

We would like to express our gratitude to the anonymous reviewers of HSP-2023 and CogSci-2024 for their invaluable feedback. The human data in this paper was collected according to Cornell IRB protocol IRB0010461 (2107010461).

\bibliographystyle{apacite}

\setlength{\bibleftmargin}{.125in}
\setlength{\bibindent}{-\bibleftmargin}

\bibliography{deplen,extra}

\end{document}